# Causal Analysis of Customer Churn Using Deep Learning

David Hason Rudd[#1,a], Huan Huo[*2,b], Guandong Xu[#3,c]

[#1&2]The University of Technology Sydney, Sydney, Australia

[#3]Advanced Analytics institute (AAi), Sydney, Australia

[a]david.hasonrudd@student.uts.edu.au

[b]huan.huo@uts.edu.au

[c]guandong.xu@uts.edu.au

*Abstract*— **Customer churn describes terminating a relationship with a business or reducing customer engagement over a specific period. Two main business marketing strategies play vital roles to increase market share dollar- value: gaining new and preserving existing customers. Customer acquisition cost can be five to six times that for customer retention, hence investing in customers with churn risk is smart. Causal analysis of the churn model can predict whether a customer will churn in the foreseeable future and assist enterprises to identify effects and possible causes for churn and subsequently use that knowledge to apply tailored incentives. This paper proposes a framework using a deep feedforward neural network for classification accompanied by a sequential pattern mining method on high-dimensional sparse data. We also propose a causal Bayesian network to predict cause probabilities that lead to customer churn. Evaluation metrics on test data confirm the XGBoost and our deep learning model outperformed previous techniques. Experimental analysis confirms that some independent causal variables representing the level of super guarantee contribution, account growth, and customer tenure were identified as confounding factors for customer churn with a high degree of belief. This paper provides a real-world customer churn analysis from current status inference to future directions in local superannuation funds.**

## I. INTRODUCTION

This Businesses rely heavily upon retaining satisfied customers in the global marketplace, which account for a large portion of their revenue. As the market becomes increasingly saturated, businesses have learned to emphasize maintaining existing clients. Although obtaining new clients was critical for initial business success, retention policies should have equal weight. Many previous studies have identified retention rate's substantial effect on the market, but clients are always able to churn away from a business, resulting in potential losses for the organization. However, customers usually offer some warning before being churned, hence churn prediction systems primarily focus on customer behavior to identify specific customers who are likely to churn out and indicate reasons for the churn. Such factors would aid marketing to develop effective retention strategies, increasing overall customer lifetime value, and assisting in growing the company's market value. For example, banks or financial organizations churn customers who close their accounts and had fewer transactions/interactions over time [1]. Churn rate,

ChurnRate = LostCustomers/InitialCustomers

is the percentage of customers a company loses in a fixed period, and represents the business pulse [2]. Voluntary churn is where a consumer wants to leave the company on their own. This could be due to dissatisfaction with product, or perhaps they feel they are not receiving that value they expected. In contrast, involuntary churn is where a customer leaves the company for unavoidable reason(s), such as payment problems due to expired account details, network issues, or inadequate cash; and incidental churn occurs due to location or financial situation changes. Deliberate churn arises due to customer desire to change innovation and price, where the most common reasons are poor service quality, non-competitive pricing, and missing customer expectations. Companies can utilize churn analysis to determine the individual user risk levels and develop appropriate focused retention initiatives.

The proposed framework applies to churn analysis for a local superannuation fund. Data preparation and analysis verified that most customer accounts were active for less than one year, for several possible reasons, including some customers being involuntary churners who were subsequently enrolled in different superannuation fund(s) operated by their employers. In contrast, voluntary churn occurs when a customer decides to cease the account for personal reasons, including perceived quality, technology, and price.

As our main contribution to this study, we built a specific churn propensity modeling work-frame to analyze customer financial behavior data, then generalized the model for any superannuation fund(s) businesses. First, we analyzed 12-month time window data to predict churners for the next six months. Since mining data from the 12-month observation window allowed us to extract the latent factors cause of churn. Then, we scaled up the model by exploiting Bayesian networks to describe how deliberate churn occurs due to multiple causes and showed that hypothesis tests on common features greatly influence prediction results. Finally, we proposed a specific causality analysis method that can be applied to other similar datasets employed in most superannuation funds.

It should be noted that this study only focuses on the performance of a DFF NN to address churn prediction on a massive and high-dimensional sparse dataset that is usually created in financial networks. We conducted pre-processing phase in the proposed model for the baseline classifiers to address the sparse and skewed data and then compared the results with the output of the proposed DFF NN. These very high dimensions data are primarily generated in subscribed-base businesses, mostly in superannuation funds due to employing interval-based features used in customer relationship management (CRM) systems.

The remainder part of the paper proceeds as follows: Section II reviews recent relevant churn studies, and Section III gives preliminary knowledge on the application of the deep learning method in churn prediction problems. Section IV discusses specific research and analysis methods employed to predict underlying reasons for churn, including problem definition and churn propensity modeling workflow. Section V analyses gathered data to address research questions: a causal analysis of churn through the financial data collected by superannuation funds and present experiments causality analysis outcomes. Finally, Section VI summarizes and concludes the paper and discusses implications from the findings for real-world applications.

## II. RELATED WORKS

### A. Customer churn prediction and causal analysis

Most previous studies focused on determining churn variables for a particular dataset rather than customer churn causation analysis.

CHAMP (Churn analysis, modeling, and prediction) is an integrated system for forecasting consumers canceling their cellular phone service [3]. Alyuda Neuro-Intelligence employs neural networks (NNs) for data mining to forecast customer churn at banks [4]. Integrating textual data using customer churn prediction (CCP) algorithms adds value [5], and combining different classifiers, e.g., gradient boost, oversampling, and contrast sequential pattern mining on single year observation windows, has been shown to be a practical strategy to deal with highly skewed data collected from superannuation funds [6]. Different churn prediction techniques have been evaluated to identify optimal approaches [7].

Hidden churn is a common problem for superannuation funds, where customer accounts become dormant once mandatory employer payments cease. Various remedies for insufficient consumer interaction have been proposed, addressing unbalanced and fully leveraged data problems, and multiple classifiers have been developed from sampled datasets [6]. Deep learning techniques can handle very large datasets compared with standard machine learning (ML) approaches, and combining deep learning and convolutional neural networks (CNNs) have successfully forecast churn [8].

Recent causal inference developments highlight fundamental alterations required to move from standard statistical analysis towards multivariate causal analysis [3]. New PC-Stable approaches effectively learn causal structures using DFS data, allowing temporal causal modeling for enormous time information datasets [9]. The directed acyclic graphs have been proposed to represent causal relationships in Bayesian networks [10]. A likelihood of churn is anticipated, as are the driving factors on banking data. Subsequently, Shah et al. [11] trained a model to produce suitable weights for features that predict whether a customer would churn. They contrasted churn definitions commonly used in business administration, marketing, IT, telecommunications, newspapers, insurance, and psychology. Many studies have described churn loss, feature engineering, and prediction models using this approach [2].

The above approaches have been variously adapted for current churn prediction methods, but most are employed for collecting data from telco, banks, and retail markets. In contrast, the proposed approach was tested using a massive financial dataset with high-dimensional sparse data. Furthermore, to our best knowledge, no previous study considered causality analysis for customer churn in superannuation companies. Motivated by the studies mentioned above, we proposed a new approach that combines deep learning and Bayesian theory to improve churn prediction modeling systems.

## III. PRELIMINARY KNOWLEDGE

Churn prediction models will provide earlier churner identification and hence assist customer intervention program development. This is basically a classification problem, i.e., to categorize each customer as a potential churner or non-churner. Several machine learning techniques for churn prediction have produced verifiable results from interpretable models, such as boosting technique, a non-parametric method, logistic regression. However, this approach is only valid predictability under circumstances where the size of the customer database is very small, varying sample size, and fails for larger datasets [7]. Therefore, we propose a deep learning (DL) algorithm to deal with massive financial data volumes since deep learning (feature transformation) differentially weights features using historical data. Deep learning algorithms follow human brain architecture using artificial neural networks (ANNs) [12]. Feed-forward NNs use output from one layer as input for the next layer with no loops between layers [13]. Therefore, we employed the deep learning feed-forward (DFF) algorithm for experiments, trading-off between reduced computational load and highest accuracy for different algorithms. General advantages for DFF NNs can be summarized as

1) superior accuracy from DL;

2) includes more variables than classical ML;

3) DL algorithms can extract patterns while avoiding blind spots from extensive customer demographics, behavioral variables, and billions of customer engagement logs.

4) reduces time-consuming feature engineering and manual financial data analysis.

We employed a DFF NN approach not only due to DL useful features, but also we only need to change the last layer(s) to update the proposed model for future data, skipping time-consuming training models. One significant weakness for deep learning is that it remains a black-box model, i.e., DL does not express uncovered patterns in the underlying data in easily understandable ways. However, we conducted experimental analyses to identify churn effects and possible causes using Bayesian networks to address NN model complexity.

## IV. METHODOLOGY

### A. Problem definition

Data mining can identify useful knowledge in terms of pattern extraction from different sources, and various feature engineering tools can extract hidden patterns from massive datasets [12], [14]. Figure 1 shows that this study included 12 datasets from a superannuation company for real-world experiments to verify the proposed model's effectiveness. We define a customer as a churner if they closed their account during the subsequent 6-month time window. Therefore, we use binary outcome for each customer [0 or 1], where 1 means the account closed and 0 that it was not close in the subsequent 6-month time window.

We applied two main inclusion criteria for account tenure and balance to satisfy model development. First, we only retained data for customers with more than 6-month account tenure; and second, we removed account balances below $1500 to improve prediction since predicting churn probability for inactive accounts has low value for superannuation funds. Figure 2 shows features in the observation window to predict which user will be churned or non-churned in the next 6-month outcome window.

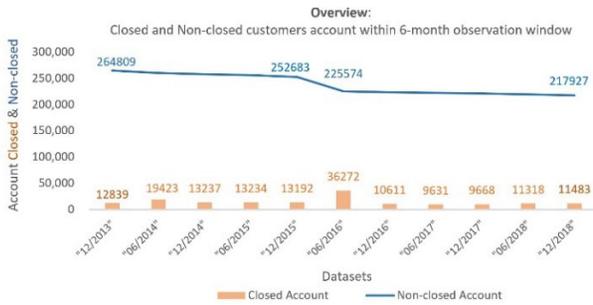

Fig. 1 Closed and non-closed customer accounts in time periods of 6-month

### B. Churn propensity modeling

We compared the proposed DFF NN algorithm approach with seven current best-practice classifiers to evaluate the proposed framework effectiveness on the datasets with a 12-month observation window and 6- month test window, as discussed in the problem definition section. The workflow of the proposed model is shown in Figure 3. Feature engineering algorithms were employed for training (80%), and test (20%) sets to derive 193 features from 124363 total examples. Binary classification generally produces severely skewed data distributions, and hence we employed the synthetic minority oversampling technique (SMOTE) [17] in pre-processing to synthesize new examples for the minority class to ensure an equal sample count (14031) for each class (1 or 0) in the training set. Experimental results confirmed improved performance for the proposed model using the sampling method.

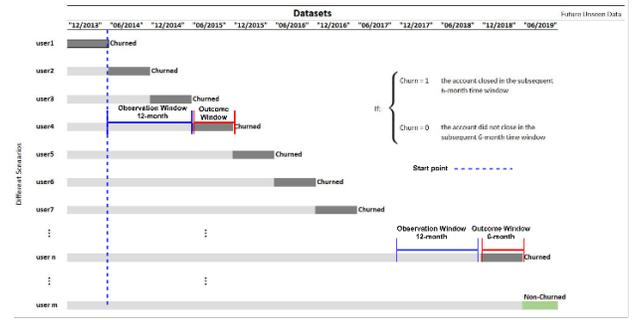

Fig. 1 Extracting feature values by sliding observation window on datasets

Furthermore, we employed a majority voting ensemble from the scikit-learn library [18] to combine predictions from multiple models, which can be helpful to improve model performance for classification tasks. Ensemble hard and soft voting were both applied on the comparison supervised models. Hard voting counts each individual classifier's votes and the majority wins. In contrast, soft voting weights each prediction by classifier importance, and the target label with the greatest sum of weighted probabilities wins [18].

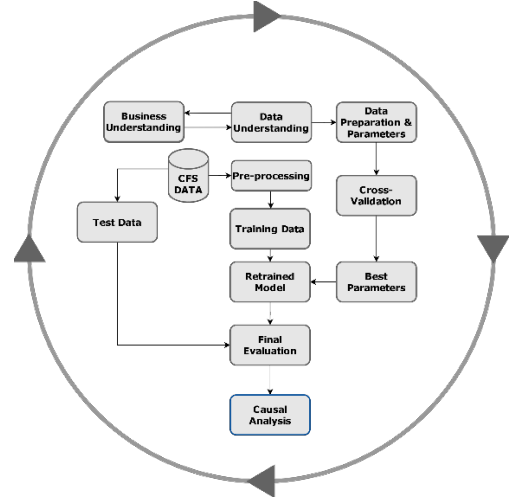

Fig. 3 Churn propensity modeling

TABLE I: NETWORK ARCHITECTURE

| Network type | DFF neural network |
| --- | --- |
| Hidden layers | 4 |
| Dense activation 1,2, and 3 | tanh, and 2 x relu |
| Dense activation 1,2, and 3 | 0.3 and 0.0 |
| Output activation function | Sigmoid |
| Learning rate | 0.000474718 |
| Epochs | 100 |
| Optimization algorithm | ADAM |

We defined a specific ANN architecture to address the research problem. Table I shows the DFF ANN's architecture optimized based on the hyperparameter tuning analysis. Finally, in the last step of the churn propensity model, we considered features that have the most predictive power on classifier outcome as causal variables. Then, we used a directed acyclic graph (DAG) to trace the latent variables and unknown parameters. Therefore, we take advantage of both Machine learning (ML) and Bayesian systems in this study. In ML, we predicted the target or independent variable by observing dependent variables, selected the most influential features to enhance prediction accuracy, and used them in our proposed causal Bayesian network. Thereinafter we employ a different approach for causal analysis. We built a dependency structure between dependent features, i.e., a causal graph, based on statistical relationships between independent features. Different methods to identify causal traces from large, labeled datasets enabled predicting more flexible causal relations [19].

## V. EXPERIMENT SETUP AND RESULTS

### A. Datasets

Experiments were conducted on 12 datasets for members holding accounts provided by a local finance company. Datasets included customer account, demographics, customer engagement, and financial data. Each dataset included approximately 250,000 examples with 88 features (71 numerical and 17 nominal). Data cleaning used criteria detailed in Section IV to improve prediction accuracy. Figure 4 visualizes customer and account tenure distribution data that are almost the same on the 12 datasets. It can be inferred that approximately 90% of customers had 5–12-month account tenure, with average tenure ≈ 7.6 months. Thus, most customers closed their accounts in less than one year. Therefore, we adopted the rational approach to only retain customer accounts that were open for more than six months, i.e., we eliminated all recently opened accounts.

### B. Evaluation metrics

Measuring model performance is essential for ML, and several techniques evaluate model effectiveness for regression and classification tasks. We used the area under the curve (AUC) and the Recall to visualize a classifier's overall performance. The AUC represents how capable the model is to discriminate classes [15], with larger AUC indicating better distinguishability between churned and non-churned customers, and AUC ≈ 1 represents good separability. AUC can be defined as [6]:

$$AUC = \frac{1}{mn} \sum_{i=1}^{n} \sum_{j=1}^{m} 1.pi > pj \quad (1)$$

where $i$ denotes to whole data points from 1 to $m$ when the churn label 1 in churned class $pi$ and $j$ denotes to all data points represent from 1 to $n$ when the churn label 0 in non-churned class. Both $pj$ and $pj$ are probabilities outcome related to each class, here 1 is an indicator when the condition $pi > pj$ is true.

### C. Correlation analysis

Relationships between variables can be measured as their correlation defined as [16]:

$$Cor(X,Y) = \frac{\sum_{i=1}^{n}(x_i - \bar{x})(y_i - \bar{y})}{\sqrt{\sum_{i=1}^{n}(x_i - \bar{x})(y_i - \bar{y})}} \quad (2)$$

We used correlation analysis to identify potentially meaningful connections between variables, selecting highly correlated relationships for subsequent causal discovery and hypothesis testing, as described in Section IV.

### D. Causality analysis experiment setup

We employed Bayesian causal graphs to encode assumptions and determine dependency levels between features, using the DoWhy python package [20]. DoWhy performs causal discovery on all potential ways to identify a desired effect based on the Bayesian causal graph model, exploiting graph-based criteria to find possible interpretation methods [20].

### E. Prediction results

Figure 5 shows experiment outcomes conducted on the most updated observation and outcome windows. XGBoost outperformed the other algorithms with AUC = 80%, whereas the proposed DFF NN and Random Forest (RF) have almost the same performance, with improved AUC by 7.5% compared with logistic regression. Thus, DFF NN outcomes were comparable with current best practice classifiers and achieved maximum prediction accuracy on test data. These findings provide a solid evidence base for exploiting DL reduced time-consuming feature engineering requiring expert knowledge on these specific financial datasets.

Descriptive approaches in statistical analysis define feature weights reflecting their contribution to push model output from its base (average output on the training dataset) to more meaningful outputs. Figure 6 shows SHAP, representing feature impacts on model output, where red features increase and blue reduce prediction outcomes relative to the base value, e.g., feature acc balance change amount reduced and sg recency increased prediction outcomes.

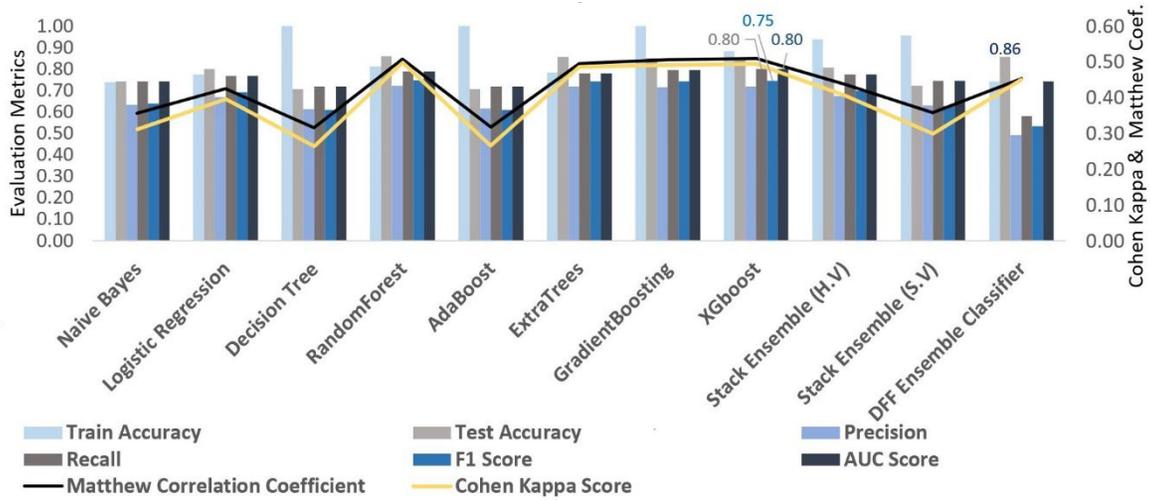

Fig. 5: Proposed DFF NN and current best-practice model

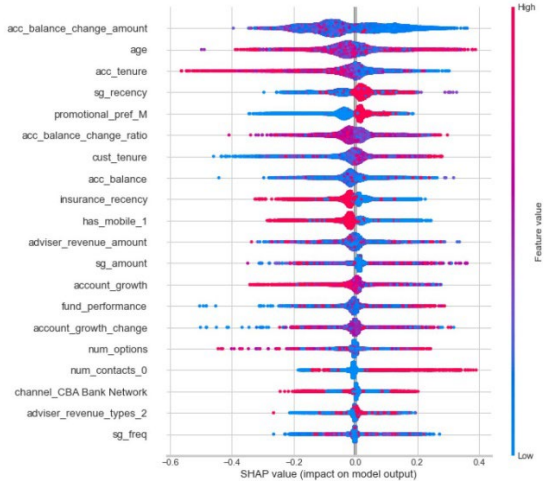

Fig. 6 SHAP visualization on prediction outcomes. The age, acc tenure, and sg recency are the most influential features pushing the prediction higher, unlike feature acc balance change amount that is pulling the model's prediction lower.

*F. Causality analysis results*

Figure 7 shows empirical results and related causal graphs based on assumptions that would affect churn as follows:

- High-limit account balance might affect the churn as users with a lower limit (account balance) might not be loyal to the superannuation funds compared to customers with a high account balance.
- The account balance change amount could affect the customer tenure. The customer tenure is often based on account balance, after all. The account balance amount itself might affect the churn. An account balance should not be below $1500 in most superannuation funds, and a low account balance generally represents an inactive customer who may be willing to churn.
- A cascade relation between account balance and gender shows that gender would affect account balance then indirectly affect churn.
- Account growth and high balance change could directly affect churn. No variation account growth indicates stopped business for most accounts with consequential increased churn propensity.

We identified causal treatment effects on churn outcome based on the initial assumptions by holding other potential effects constant while changing the target treatment. Linear regression estimation indicates that effect = −0.026298 corresponds to churn probability reducing by ≈ 2.5% when the customer has higher account tenure. To test our assumption so that if the assumption is correct, the new estimation effect should not significantly change. Therefore, we applied random common cause (RCC) [20] to refute obtained estimates by adding independent random variable res random as a common cause in the dataset. The outcome from the refuting method = −0.026299, almost identical to the estimation result. Thus, we can confirm that the assumption was correct that high account tenure was a causal feature for churn outcome.

The treatment's causal effect on the outcome is based on the change in the value of the treatment variable. How strong the effect is a matter of statistical estimation. There are many methods for the statistical estimation of the causal effect. In this study, we used the" Propensity Score-Based Inverse Weighting" method [21], and concluded the estimations and churn probability results in Table II. For instance, the mean estimation for variable sg recency is ∼ 0.15, where it is equivalent to saying that the probability of churn is increased by ∼ 15% when the customer has higher days since the last day of super guarantee SG contribution. The mean estimates of ∼ 0.03 for the account growth variable can be concluded that churn probability increases by ∼ 3% when the customer has a negative account growth rate. The python implementation for the proposed framework and result visualization, including hyper-parameter sensitivity analysis of DFF NN in hiplot, and our causal analysis in detail, is available in GitHub (https:

//github.com/DavidHason/Causal_Analysis) to simplify reproducing and improving this study experiment results.

## VI. CONCLUSION

Losing customers is inevitable for most businesses, but churn can be managed at acceptable levels by investing in customers with churn risk. A new churn propensity model was designed and integrated with the Bayesian causal network. Unbalanced churned and non-churned classes were leveled in pre-processing, and then ac- curacy was compared between the proposed DFF NN approach and ten current best-practice churn prediction classifiers. Although XGBoost achieved superior AUC, DFF NN obtained comparable AUC with the highest accuracy of all considered models on test data.

We analyzed possible customer churn causes for a particular financial dataset created at superannuation fund(s) corporations. Causal analysis results confirmed variables representing recent SG contribution, annual report preference changed, account growth, and the balance amount was identified as confounding factors for customer churn with a high degree of belief. Churn rate can be reduced by ∼ 3% for customers with active accounts> 1year, consistent with expert knowledge.

A natural progression of this work is extending pattern mining techniques with smaller outcome windows that should be investigated to obtain more efficient prediction results in future studies. Also, different methods to identify causes of churn based on counterfactual causal analysis should be investigated.

## ACKNOWLEDGMENT

This work is partially supported by the Australian Research Council (ARC) under Grant No. DP200101374 and LP170100891.